\documentclass{article} 
\usepackage[nonatbib,final]{nips_2018}

\usepackage[T1]{fontenc}    
\usepackage{hyperref}       
\usepackage{url}            
\usepackage{booktabs}       
\usepackage{amsfonts}       
\usepackage{nicefrac}       
\usepackage{microtype}      

\usepackage{mathtools}
\usepackage{subfig}

\usepackage{float}
\usepackage{epsfig}
\usepackage{amsmath}
\usepackage{amsthm}
\usepackage{amssymb}

\usepackage{algorithm}
\usepackage{algpseudocode}

\usepackage{algpseudocode}
\usepackage{framed}
\usepackage{longtable}
\usepackage{epstopdf}
\usepackage{url}
\usepackage[utf8x]{inputenc}
\usepackage{amsfonts}
\usepackage{amsmath}
\usepackage{MnSymbol}
\usepackage{mathrsfs}
\usepackage{graphicx}
\usepackage{caption}

\numberwithin{equation}{section}

\newcommand{\eq}{\begin{equation*}}
\newcommand{\en}{\end{equation*}}
\newcommand{\eqa}{\begin{eqnarray*}}
\newcommand{\ena}{\end{eqnarray*}}
\newcommand{\eqn}{\begin{equation}}
\newcommand{\enn}{\end{equation}}
\newcommand{\be}{\begin{equation}}
\newcommand{\ee}{\end{equation}}
\newcommand{\eqan}{\begin{eqnarray}}
\newcommand{\enan}{\end{eqnarray}}

\newcommand{\nn}{\nonumber}

\newcommand{\cc}{ {\bf c} }

\newcommand{\x}{ {\bf x} }

\usepackage[noabbrev,capitalize,nameinlink]{cleveref}

\usepackage{amsfonts}

\newcommand{\pmat}{\begin{pmatrix}}
\newcommand{\pman}{\end{pmatrix}}

\graphicspath{{../../figures/}{../figures/} }

\usepackage{authblk}

\begin{document} 
\title{Amortized  Bayesian inference for clustering models} 
\date{}

\author{Ari Pakman}
\author{Liam Paninski}
\affil{	Department of Statistics \\
	Center for Theoretical Neuroscience \\
	Grossman Center for the Statistics of Mind \\
	Columbia University
	}
\renewcommand\Authands{ and }

\maketitle

\begin{abstract} 
We develop methods for efficient amortized approximate Bayesian inference over posterior distributions of probabilistic clustering models, 
such as Dirichlet process mixture models. The approach  is based on mapping distributed, symmetry-invariant representations of cluster arrangements into conditional probabilities. 
The method parallelizes easily, yields iid samples from the approximate posterior of cluster assignments
with the same computational cost of a single Gibbs sampler sweep, 
and can easily be applied to both conjugate and non-conjugate models,
as training only requires samples from the generative model.
\end{abstract} 

\section{Introduction}
Unsupervised clustering is a key tool in many areas of statistics and machine learning, 
and analyses based on probabilistic generative models are crucial 
whenever there is irreducible uncertainty about the 
number of clusters and their members.

Popular posterior inference methods 
in these models fall into two broad classes. 
On the one hand, MCMC methods~\cite{neal2000markov, jain2004split,jain2007splitting} 
are asymptotically accurate but time-consuming, with convergence that is difficult to assess. Models whose  likelihood and prior are non-conjugate
are particularly challenging, since in these cases the model parameters cannot be marginalized and
must be kept as part of the state of the Markov chain. 
On the other hand, variational methods~\cite{Blei:2004:VMD,kurihara2007collapsed,hughes2015reliable} are typically much faster but do not come with accuracy guarantees.

In this work we propose a novel approximate 
amortized approach,  
based on training  neural networks to 
map distributed, symmetry-invariant representations of cluster arrangements into conditional probabilities. The method can be applied to both conjugate and non-conjugate models, and 
after training the network with samples from a particular generative model, we can 
obtain independent,
GPU-parallelizable, approximate posterior samples of cluster 
assignments for any new set of observations of arbitrary size, with no need for expensive MCMC steps.

\section{The Neural Clustering Process}
Probabilistic models for clustering~\cite{mclachlan1988mixture} 
introduce random variables $c_i$ denoting  the cluster number to which 
the data point $x_i$ is assigned, and assume a generating process 
of the form
\eqan
\label{gen1}
c_1 \ldots c_N &\sim& p(c_{1},\ldots, c_{N}) 
\\
\mu_k &\sim& p(\mu_k) \quad k=1 \ldots K
\\
x_i &\sim& p(x_i|\mu_{c_{i}}) \quad i=1 \ldots N
\label{gen3}
\enan
Here $K$ is the number of distinct values among the $c_i$'s, $\mu_k$ denotes a parameter vector controlling the distribution of the $k$-th cluster,
and $p(c_{1:N})$ is assumed to be exchangeable.
Examples of this setting include 
Mixtures of Finite Mixtures~\cite{miller2018mixture} and
many Bayesian nonparametric models, such as Dirichlet process 
mixture models (DPMM) (see~\cite{npb_review} for a recent overview).

Given  $N$ data points $\mathbf{ x} = \{x_i\}$, 
we are interested in  sampling the $c_i$'s, using a decomposition
\eqan 
p(c_{1:N}|\mathbf{x}) 
=  p(c_1| \mathbf{ x} ) p(c_2|c_1,\mathbf{ x} ) \ldots p(c_N|c_{1:N-1}, \mathbf{ x}).
\label{joint}
\enan 
Note that $p(c_1| \mathbf{ x} )=1$, since the first data point is always assigned 
to the first cluster. To motivate our approach, it is useful to consider the joint distribution
of the assignments of the first~$n$ data points,
\eqan 
p(c_1, \ldots, c_n| \mathbf{ x}) \,.
\label{joint_n}
\enan 
We are interested in representations of $\mathbf{ x}$ that keep  the symmetries of (\ref{joint_n}):

\begin{itemize}

	\item {\bf Permutations within a cluster: }	
	(\ref{joint_n}) is invariant under permutations of $x_i$'s belonging to the same cluster.
	If there are $K$ clusters, each of them can be represented by 
		\eqan 
		H_k= \sum_{i : c_{i}=k} h(x_i)  \qquad k = 1\ldots K\,,
		\enan 
	where $h:\mathbb{R}^{d_x} \rightarrow \mathbb{R}^{d_h}$ is a function we will learn from data.
	This type of encoding has been shown in~\cite{deep_sets}
	to be  	necessary to represent functions with permutation symmetries.

	\item {\bf Permutations between clusters: }
	(\ref{joint_n}) is invariant under permutations of the cluster labels. 		
	In terms of the within-cluster invariants~$H_k$, this symmetry can be captured by 	
	\eqan 
	G= \sum_{k =1}^K g(H_k) ,
	\label{G_def}
	\enan 			
	where $g:\mathbb{R}^{d_h} \rightarrow \mathbb{R}^{d_g}$.
	
		\item {\bf Permutations of the unassigned data points: }
		(\ref{joint_n}) is also invariant under permutations of the $N-n$ 
		unassigned data points. 
		This can be captured by 
		\eqan 
		Q = \sum_{i=n+1}^{N}  h(x_i) .
		\enan 		
\end{itemize} 
Note that $G$ and $Q$ provide fixed-dimensional, 
symmetry-invariant representations of all the assigned and non-assigned data points, respectively,
for any number of $N$ data points and
$K$ clusters.  Consider now the conditional distribution that interests us,
\eqan
p(c_n|c_{1:n-1}, \mathbf{ x}) 
= \frac{p(c_1 \ldots c_n| \mathbf{ x})}
{ \displaystyle \sum_{c_n'=1}^{K+1}  p(c_1 \ldots c_n'| \mathbf{ x})}.
\label{conditional}
\enan 
Assuming $K$ different values in $c_{1:n-1}$,
then $c_n$ can take $K+1$ values, corresponding to $x_n$ joining any of the $K$ existing clusters, or forming its own new cluster.
Let us denote by $G_k$ the value of (\ref{G_def}) for each of these $K+1$ configurations. 
In terms of the $G_k$'s and $Q$, we propose to model (\ref{conditional}) as
\eqan 
p_{\theta}(c_n=k|c_{1:n-1}, \mathbf{ x}) = \frac{ e^{f(G_k,Q, h_{n} )} }
{  \sum_{k'=1}^{K+1} e^{f(G_{k'},Q, h_{n})}       } 
\label{ncp}
\enan
for $k = 1\ldots K+1$, 
where $h_{n}=h(x_n)$ and $\theta$ denotes all the parameters in the functions $h,g$ and $f$, that will be represented with neural networks. 
Note that this expression preserves the symmetries of the numerator and denominator in the rhs of (\ref{conditional}).
By storing and updating $H_k$ and $G$ for successive values of~$n$, the computational 
cost of a full sample of $c_{1:N}$ is $O(NK)$, 
the same of a full Gibbs sweep. See~\cref{algo} for details; we term this approach the Neural Clustering Process (NCP).


\subsection{Global permutation symmetry}
\label{sec:global}
There is yet another symmetry present in the lhs of (\ref{joint})
that is not evident in the rhs: a global simultaneous permutation of the $c_i$'s. 
If our model learns the correct form for the conditional probabilities, 
this symmetry should be (approximately) satisfied.  We monitor this 
symmetry during training.

\section{Learning}
In order to learn the parameters $\theta$, we use  stochastic 
gradient descent to minimize the expected negative log-likelihood,
\eqan 
L(\theta) =
- \mathbb{E}_{p(N)}    \mathbb{E}_{p(c_{1},\ldots, c_{N},\mathbf{ x})}
\mathbb{E}_{p(\pi)}
\left[  \sum_{n=2}^{N} \log p_{\theta}(c_{\pi_n} |c_{\pi_1:\pi_{n-1}},\mathbf{ x})
\right] ,
\label{expected_nll}
\enan 
where $p(N)$ and $p(\pi)$ 
are  uniform  over a  range of integers and over $N$-permutations,
and samples from $p(c_{1},\ldots, c_{N},\mathbf{ x})$ are obtained from the generative model
(\ref{gen1})-(\ref{gen3}), 
irrespective of the model 
being conjugate. 
In Appendix~\ref{app:rao-blackwell} we show that 
(\ref{expected_nll}) 
can be partially Rao-Blackwellized.



\begin{algorithm}[t]
	\caption{$O(NK)$ Neural Clustering Process Sampling }
	\label{algo}
	\begin{algorithmic}[1]

		\State $h_i \gets h(x_i)   \qquad i=1 \dots N$
		\State $Q \gets \sum_{i=2}^N h_i$  \Comment{Initialize unassigned set }
		\State $H_1 \gets h_1$      \Comment{Create first cluster with $x_1$}
		\State $G \gets g(H_1)$
		\State $K \gets 1$, $c_1 \gets 1$
		
		\For{$n\gets 2\ldots N$}
		\State $Q \gets Q - h_n$  \Comment{Remove $x_n$ from unassigned set}
		\State $H_{K+1} \gets 0$ \Comment{We define $g(0)=0$}
		\For{$k\gets 1\ldots K+1$}  
		\State $G \gets G +g(H_k + h_n) - g(H_k)$ \Comment{Add $x_n$}
		\State $p_{k} \gets e^{f(G,Q,h_n)}$ 
		\State $G \gets G - g(H_k + h_n) + g(H_k)$ \Comment{Remove $x_n$}
		\EndFor
		\State $p_k \gets p_k/\sum_{k'=1}^{K+1}p_{k'}$   \Comment{Normalize probabilities }
		\State $c_{n} \sim p_k$   \Comment{Sample assignment for $x_n$ }
		\If{$c_{n} = K+1$}
		\State $K \gets K+1$
		\EndIf 
		\State $G \gets G-g(H_{c_{n}}) + g(H_{c_{n}} + h_{n})$  \Comment{Add point $x_{n}$}
		\State $H_{c_{n}} \gets H_{c_{n}} + h_{n}$
		\EndFor
		\State \Return $c_1 \ldots c_N$			
	\end{algorithmic}
	
\end{algorithm}

\section{Related work}
The work~\cite{du2010clustering} provides an overview of 
deterministic clustering based on neural networks, 
and~\cite{Pehlevan226746} proposes a biologically inspired network 
for online clustering. Our work differs from previous approaches
in its use of neural networks to explicitly approximate fully Bayesian inference in a probabilistic generative clustering model.
Similar amortized approaches to Bayesian inference
have been explored in Bayesian networks~\cite{stuhlmuller2013learning}, 
sequential Monte Carlo~\cite{paige2016inference}, 
probabilistic programming~\cite{ritchie2016deep,le2016inference}  and particle tracking~\cite{pmlr-v80-sun18b}.
The representation of a set via a sum (or mean) of encoding vectors was also used 
in~\cite{deep_sets,edwards2016towards,  garnelo2018conditional,garnelo2018neural}.

\section{Results}
In this section we present examples of NCP clustering.
The functions $g$ and $f$ 
have the same neural architecture in all cases, and for different data types we only change the encoding function $h$. More details 
are in Appendix~\ref{app:details},
where we also show that during training the variance of the joint likelihood (\ref{joint}) 
for different orderings of the data points drops to negligible values.

\begin{figure}[t]
	\begin{center}
		\includegraphics[width=\textwidth,height=.3\textwidth]{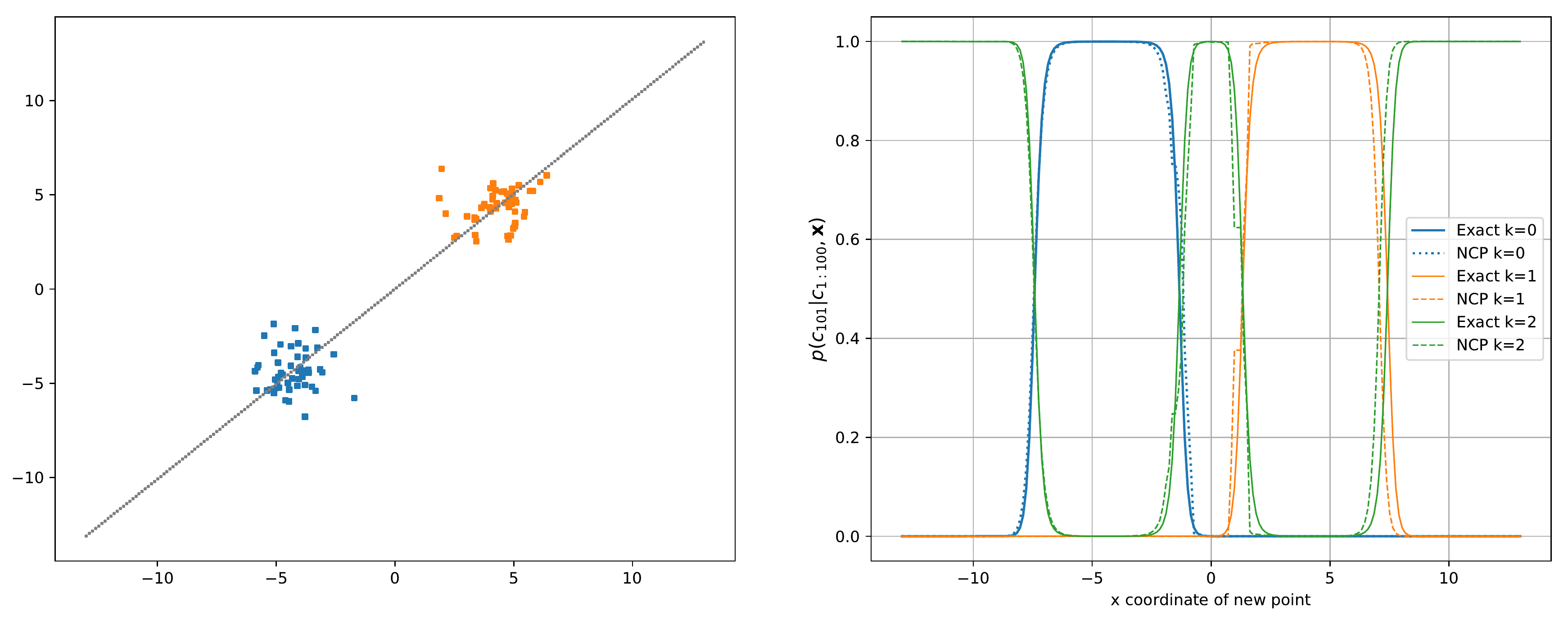}
		\includegraphics[width=\textwidth]{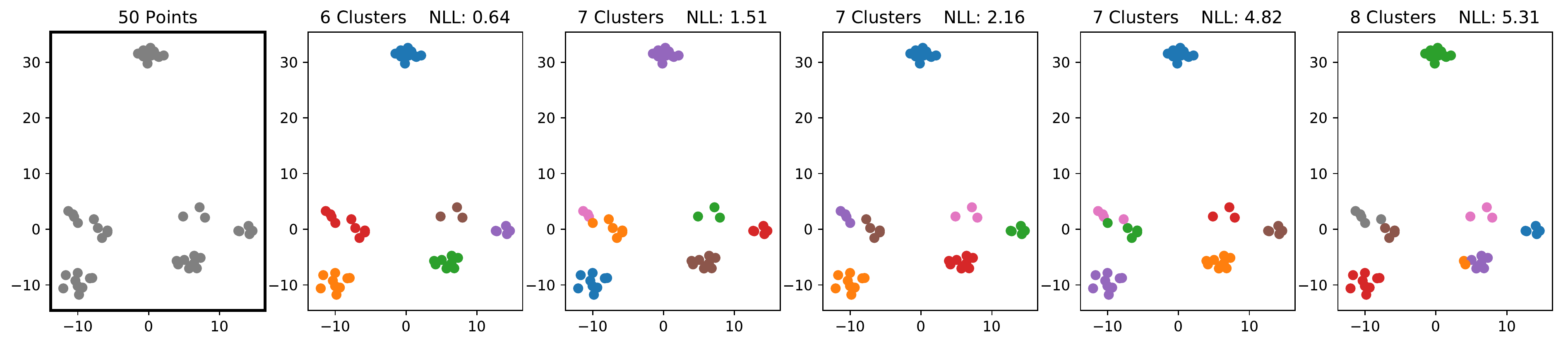}		
	\end{center}
	\caption{
		{\small 
		{\bf NCP vs. exact posteriors.} 
		{\it Upper left:} Two 2D clusters of 50 points each ($k=0,1$) and a line over possible locations of a 101st last point. 
		{\it Upper right:} Assuming a DPMM (here with $\alpha=0.7$, and 2D Gaussian observations with  unit variance and mean with a prior $N(0,\sigma_{\mu}=10 \times \mathbf{1}_2)$), the posterior $p(c_{101}|c_{1:100}, \mathbf{ x})$ 
		can be computed exactly, and we compare it to the NCP estimate as a function of 
		the horizontal coordinate of $x_{101}$, as this point moves over the gray line on the upper left panel.
		{\it Lower:} Five samples from the posterior of the same 2D model, given the observations in the leftmost panel.
		In each sample, the order of the particles was randomly shuffled.
		Note that the posterior samples are reasonable, and less-reasonable samples are assigned 
		higher negative log-likelihood (NLL) values by the NCP. (Best seen in color.)
 }
	}
\label{fig:exact_vs_ncp}
\end{figure}

Figure~\ref{fig:exact_vs_ncp} shows results for a DPMM of 2D conjugate Gaussians.
In particular, we compare the estimated assignment probabilities for a last observation of a set, $c_N$,
against their exact values, which are computable for conjugate models, showing excellent agreement.

\begin{figure}[h]
	\begin{center}		
		\fbox{\includegraphics[width=.98\textwidth]{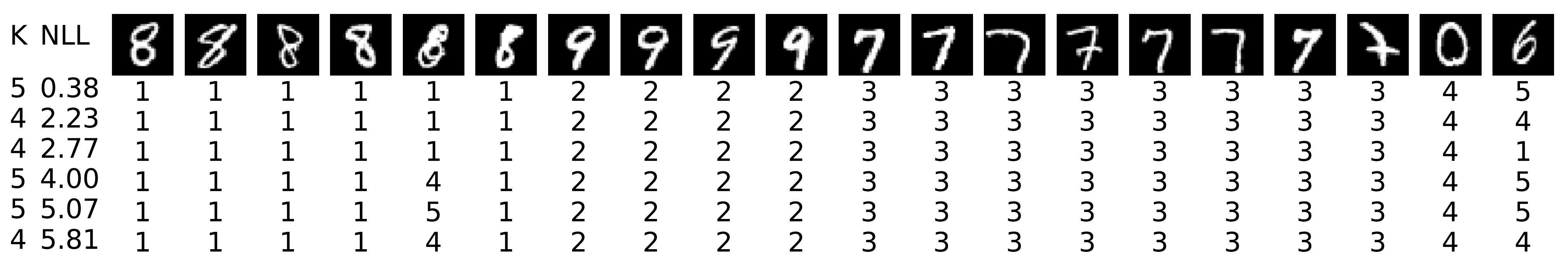}}
	\end{center}
	\caption{ 
		{\small 
			{\bf Clustering of MNIST data.} The generative model 
			is a DPMM with concentration parameter $\alpha=0.7$ and a uniform discrete base measure over
			the $10$ labels. Conditioned on a label, observations are sampled uniformly from the MNIST training set. The figure shows $N=20$ observations, generated similarly from the MNIST test set.
			The six rows below the observations 
			show six samples of $c_{1:20}$ 
			from the NCP posterior of these~20 images.
			Most samples from the NCP yield
			the first row of assignments, which has very low
			negative-loglikelihood (NLL) and 
			is consistent  with 
			the true labels. The next five rows 
			correspond to more rare samples
			from the NCP, with higher NLL, 
			each capturing some ambiguity 
			suggested by the form of particular digits. 
			In this case we drew 39 samples: 34 corresponding to the first row, and one to each of the next five rows.
		}}
	\label{fig:mnist}		
	\end{figure}

Figure~\ref{fig:mnist} shows results for a
DPMM over the empirical distribution of $28\times 28$-pixel handwritten digits from the MNIST dataset.
In this case the generative model has no analytical expression. The results show that the 
NCP samples correctly capture the label ambiguity of some of the digits.

\section{Outlook}
We have introduced a new approach to sample from (approximate) posterior distributions of probabilistic  clustering models. Our first results show reasonable agreement with Gibbs sampling, with major improvements in speed and model flexibility. 

\newpage
\onecolumn
\appendix

\section{Details of the examples}				
\label{app:details}

We implemented the functions $g$ and $f$ as  six-layered MLPs with PReLU non-linearities~\cite{xu2015empirical}, with $128$ neurons in each layer,
and final layers of dimensions $d_g=512$ for $g$ and $1$ for $f$.
We used stochastic gradient descent with ADAM~\cite{kingma2014adam},
with a step-size of $10^{-4}$ for the first 1000 iterations,
and $10^{-5}$ afterwards.
The number of Monte Carlo samples from (\ref{expected_nll}) 
in each mini-batch were: 1 for $p(N)$, 8 for $p(\pi)$,
1 for $p(c_{1:N})$ and 48 for $p(\mu_k)$ and $p(\mathbf{x}|\mu)$. 


\subsection{Low-dimensional conjugate Gaussian models} 
The generative model for the examples in Figure~\ref{fig:exact_vs_ncp}
is
\eqan 
N &\sim& \textrm{Uniform} [5,100]
\\
c_1 \ldots c_N &\sim& \textrm{DPMM}(\alpha) 
\\
\mu_k &\sim& N(0,\sigma_{\mu}^2  \mathbf{1}_2)  \quad k=1 \ldots K
\\
x_i &\sim& N(\mu_{c_{i}}, \sigma^2  \mathbf{1}_2) \quad i=1 \ldots N
\enan 
with $\alpha = 0.7$, $\sigma_{\mu}=10$, $\sigma=1$, and $d_x=2$.
The encoding function $h(x)$ is a five-layered MLPs with PReLU non-linearities, 
with $128$ neurons in the inner layers and a last layer with $d_h=256$ neurons.

\subsection{High-dimensional MNIST data} 
The generative model for the example in Figure~\ref{fig:mnist} is
\eqan 
N &\sim& \textrm{Uniform} [5,100]
\\
c_1 \ldots c_N &\sim& \textrm{DPMM}(\alpha) 
\\
l_k &\sim& \textrm{Uniform} [0,9]  \quad k=1 \ldots K
\\
x_i &\sim& \textrm{Uniform} [\textrm{MNIST digits with label } l_{c_i}] 
\quad i=1 \ldots N
\enan 
with $\alpha = 0.7$, $d_x=28\times28$.
The architecture for $h(x)$
was: two layers of [convolutional + maxpool + ReLU]
followed by [fully connected(256) + ReLU + fully connected($d_h$)], with $d_h=256$.

\subsection{Invariance under global permutations} 
\begin{figure}[t!]
	\begin{center}		
		\hspace{-1cm}\includegraphics[width=\textwidth]{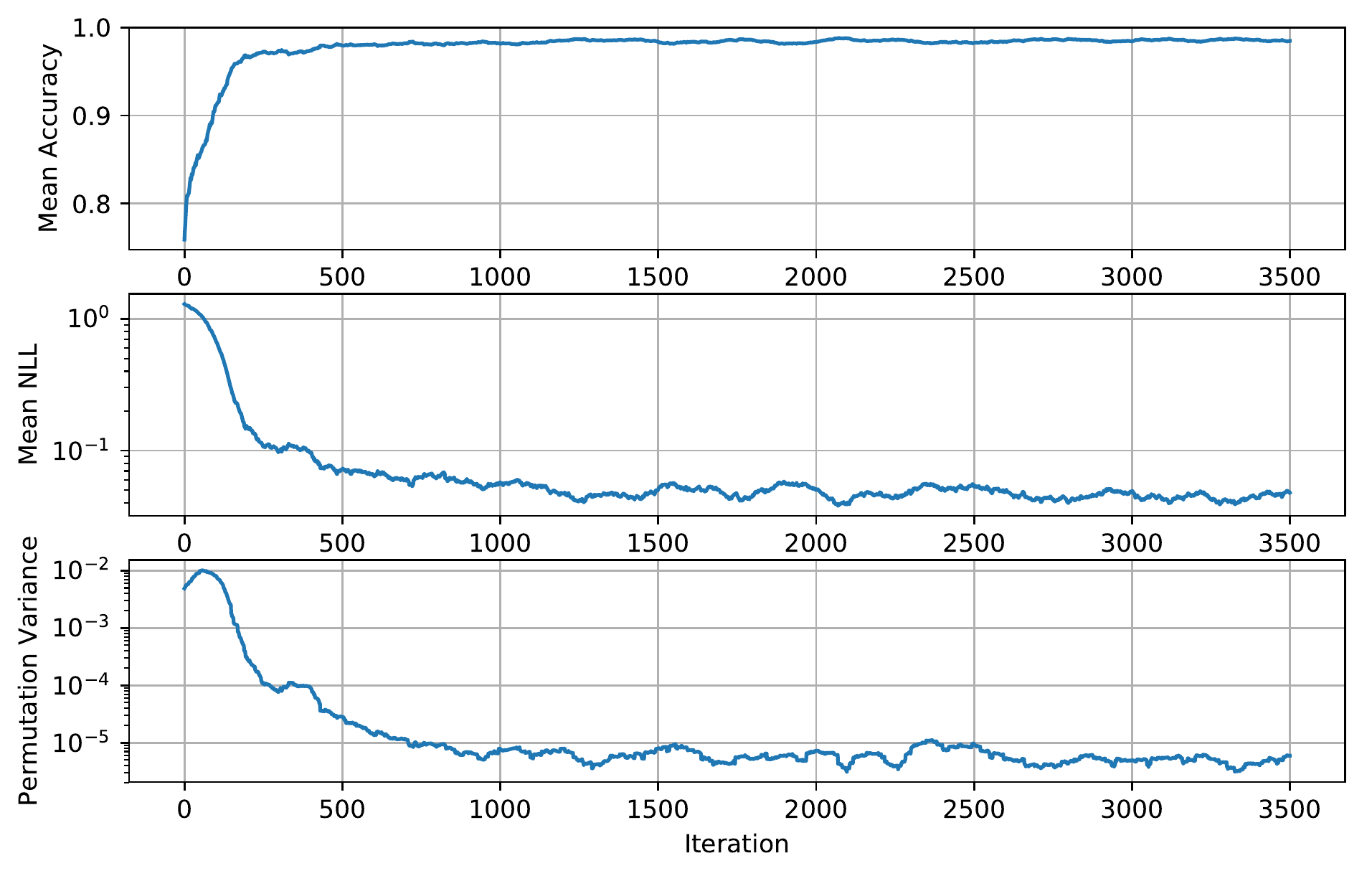}		
	\end{center}
	\caption{All the curves correspond to training of the 2D model described in Figure~\ref{fig:exact_vs_ncp}. To ease the visualization, 
			we show the averages of a sliding window of 100 previous iterations. 
			{\it Top:} `Accuracy' measures the fraction 
			of the cluster labels in the training samples that agree with the 
			the maximum of the learned categorical distribution. 
			{\it Center:} Mean negative log-likelihood in logarithmic scale.
			{\it Bottom:} Variance of the joint log likelihood under 
			global permutations, estimated from 8 random permutations,         
			in logarithmic scale. 	
		}
	\label{fig:train_avgs}
	\end{figure}

As mentioned in Section~\ref{sec:global}, if the conditional probabilities (\ref{conditional}) are learned correctly, 
invariance of the joint probability (\ref{joint}) under global permutations should hold. 
Figure~\ref{fig:train_avgs}
shows estimates of the variance of the  
joint probability under permutations as learning progresses, showing that 
it diminishes to negligible values.

\section{Importance Sampling}
\begin{figure}[t!]
	\begin{center}		
		\includegraphics[width=\textwidth, height=4.5in]{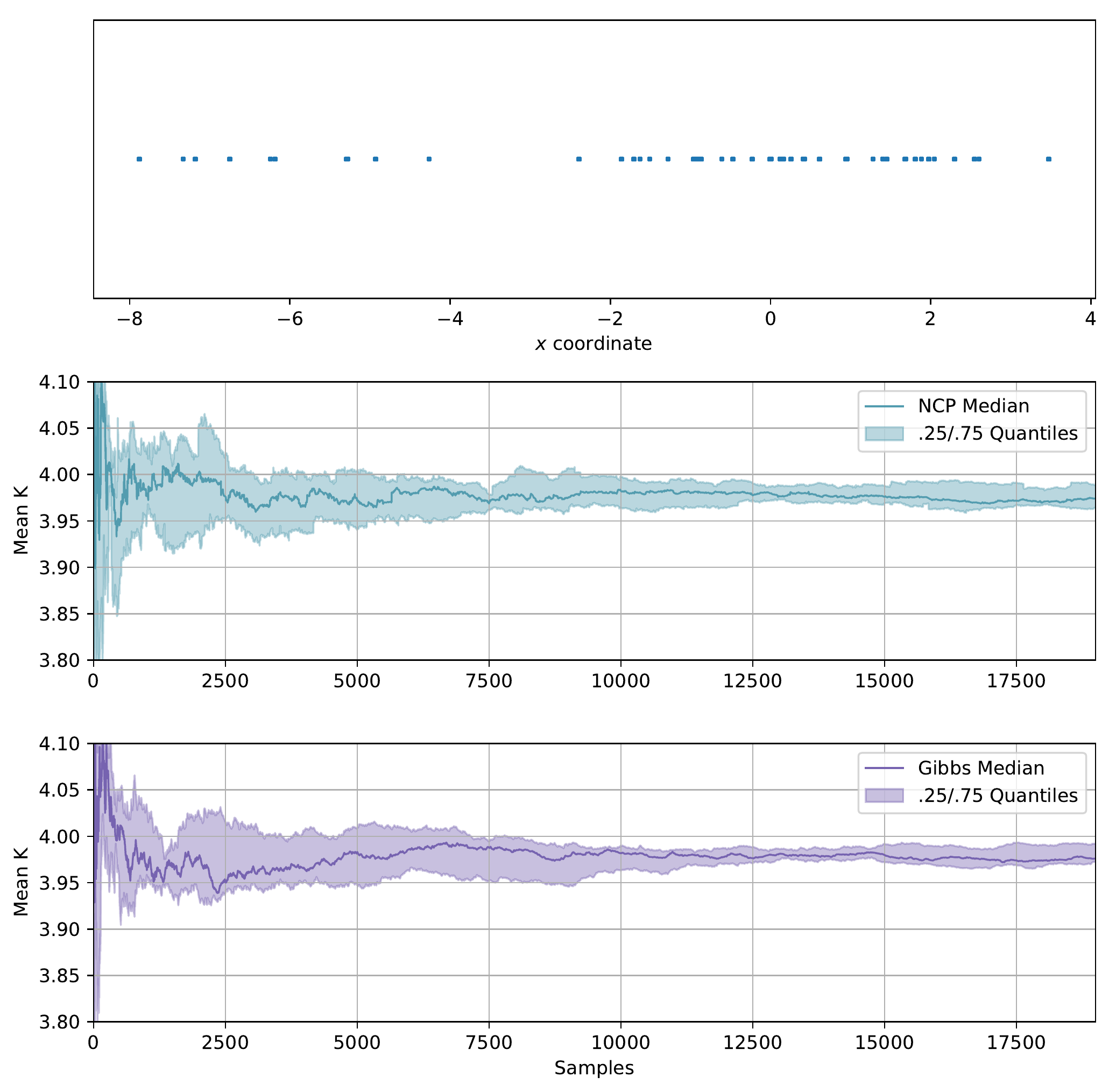}
	\end{center}
	\caption{ 			{\bf Gibbs vs NCP importance sampling.} 
	We assume a 1D generative model similar to the 2D model
	of Figure~\ref{fig:exact_vs_ncp}, 
		and compute the mean of the number of clusters $K$ as a function of the number of samples, for the dataset of $N=50$ observations in the upper panel.
			The figures show medians and $0.25/0.75$ quantiles for eight repetitions. 
			In this simple example the variance of Gibbs and NCP are comparable, but the average 
			CPU/GPU running time was 184 secs. for each NCP run of 20,000 samples, and 1969 secs. 			for  each Gibbs run (with additional 1000 burn-in samples).
			The time advantage of NCP is due to the fact that since all samples
			are iid, NCP can be massively parallelized over GPUs, while in naive implementations of the Gibbs sampler 
			the samples must be obtained sequentially.
		}
	\label{fig:importance_samples}
	\end{figure}

Samples from the NCP can be used either as approximate samples from the posterior, or 
as high-quality importance samples.  (Alternatively, we could use samples from the NCP to seed an exact MCMC sampler; we have not yet explored this direction systematically.) In the latter case, the expectation of a function $r(\cc)$ 
is given by 
\eqan 
\mathbb{E}_{p(\cc|\x)}\left[ r(\cc) \right] \simeq 
\frac{\sum_{s=1}^S \frac{p(\cc_s,\x)}{p_{\theta}(\cc_s|\x)} r(\cc_s)    }
{\sum_{s=1}^S \frac{p(\cc_s,\x)}	{ p_{\theta}(\cc_s|\x)}}
\enan 
where each $\cc_s$ is a sample from $p_{\theta}(\cc|\x)$.
Figure~\ref{fig:importance_samples} shows a comparison between 
an expectation obtained from Gibbs samples vs importance NCP samples. 

\section{Rao-Blackwellization}
\label{app:rao-blackwell}
With some more computational effort, it is possible to partially Rao-Blackwellize the expectation 
in~(\ref{expected_nll}) and reduce its variance. 
\subsection{Conjugate Models}
For given $N$ and $\x$,  a generic term in (\ref{expected_nll}) can be written is
\eqan 
\sum_{\cc} p(\cc|\x) \log p_{\theta}(c_{n} |c_{1:n-1},\mathbf{ x})
\nn
&=&
\sum_{\cc} p(c_{n:N}|c_{1:n-1}, \x)  p(c_{1:n-1}|\x)  \log 
p_{\theta}(c_{n} |c_{1:n-1},\mathbf{ x})
\\
&\simeq&  
\sum_{c_{n:N}} p(c_{n:N}|c_{1:n-1}, \x)   
\log p_{\theta}(c_{n} |c_{1:n-1},\mathbf{ x})
\label{sample_cs}
\\
&=& 
\sum_{c_{n}} p(c_{n}|c_{1:n-1}, \x)   
\log p_{\theta}(c_{n} |c_{1:n-1},\mathbf{ x})
\label{sum_cs}
\enan 
where we took here $\pi_i = i$ to simplify the notation. 
In (\ref{sample_cs}) we replaced the expectation under~$p(c_{1:n-1}|\x)$
with a sample of $c_{1:n-1}$, and in (\ref{sum_cs}) we summed over $c_{n+1:N}$.
The expectation in (\ref{sum_cs}) has lower variance than  
using a sample of $c_n$ instead. 

If there are $K$ different values in $c_{1:n-1}$, we need to compute 
about $(K+1)(K+2)\ldots (K+N-n+1)$ values for~$p(c_{n:N}|c_{1:n-1}, \x)$
in (\ref{sample_cs}),
corresponding to all the values the set $c_{n:N}$ can take.
Each of these can be computed by evaluating 
\eqan 
p(c_{n:N}|c_{1:n-1}, \x)    &\varpropto& p(\cc) p(\x|\cc)
\enan
with $c_{1:n-1}$ fixed, and then normalizing. 
Below we present an example of this computation. 

Moreover, after computing $p(\cc) p(\x|\cc)$ for fixed $c_{1:n-1}$ and all $c_{n:N}$, we can similarly Rao-Blackwellize 
all the other $N-n$ terms with $p(c_{n+1}|c_{1:n}, \x), \ldots ,
p(c_{N}|c_{1:N-1}, \x)$. Each of these distributions can be obtained 
from our original evaluation of $p(\cc) p(\x|\cc)$, by fixing 
the conditioning $c's$ and summing over the others. 

{\bf Example:} DPMM  with 1D Gaussian likelihood and Gaussian prior for the mean

The observation model is 
\eqan 
p(\mu|\lambda) &=& N(0, \sigma^2_{\mu}=\lambda^2)
\\
p(x|\mu, \sigma) &=& N(\mu, \sigma^2_{x}=\sigma^2)
\enan 
with $\lambda$ and $\sigma$ fixed. In this case we get
\eqan 
p(\x|c) &=& \prod_{k=1}^K 
\int d \mu_k N(\mu_k|0,\lambda^2) \prod_{i:c_i=k} N(x_i|\mu_k,\sigma^2)
\\
&=& \prod_{k=1}^K  \frac{\sigma_k}{\lambda} 
\exp \left(\frac{\sigma_k^2 (\sum_{i_k} x_{i_k} )^2}{2 \sigma^4 } \right) 
\exp \left(-\frac{\sum_{i_k} x^2_{i_k}}{2 \sigma^2} \right)
\enan 
where $\{ i_k \}= \{ i: c_i=k\}$ 
and $\sigma_k^{-2} = \lambda^{-2} + n_k \sigma^{-2}$, with $n_k = |i_k|$,
and 
\eqan 
p(c_{1:N}) = \frac{\alpha^{K_N}  
	\prod_{k=1}^{K_N} (n_k-1)!}{\prod_{i=1}^{N}(i-1+\alpha)}
\enan 
with $\alpha$ the Dirichlet process concentration parameter.

\subsection{Nonconjugate Case}
This case is similar, using
\eqan 
\nn
\sum_{\cc} p(\cc|\x) \log p_{\theta}(c_{n} |c_{1:n-1},\mathbf{ x})
&=&
\sum_{\cc} \int \! d\mu \, p(\cc, \mu|\x) \log p_{\theta}(c_{n} |c_{1:n-1},\mathbf{ x})
\\
\nn
&=& \sum_{\cc} \int \! d\mu \, p(c_{n:N}|c_{1:n-1}, \mu, \x)  
p(c_{1:n-1}, \mu |\x)  
\log p_{\theta}(c_{n} |c_{1:n-1},\mathbf{ x})
\\
&\simeq&  
\sum_{c_{n:N}} p(c_{n:N}|c_{1:n-1}, \mu, \x)   
\log p_{\theta}(c_{n} |c_{1:n-1},\mathbf{ x})
\label{sample_cs_nc}
\\
&=& 
\sum_{c_{n}} p(c_{n}|c_{1:n-1}, \mu, \x)   
\log p_{\theta}(c_{n} |c_{1:n-1},\mathbf{ x})
\label{sum_cs_nc}
\enan 
where now in (\ref{sample_cs_nc}) 
we replaced the expectation under $p(c_{1:n-1}, \mu|\x)$
with  samples of $c_{1:n-1}, \mu$. 
In this case we need to evaluate
\eqan 
p(c_{N-r+1:N}|\mu, c_{1:N-r}, \x) 
& \propto &
p(c_{N-r+1:N}, \x|\mu, c_{1:N-r})
\\
&=& p(c_{N-r+1:N}| c_{1:N-r})
p( \x|\mu, c_{1:N} ) \,.
\enan

\subsection*{Acknowledgments}

This work was supported by the Simons Foundation, the DARPA NESD program, and by ONR N00014-17-1-2843.

\clearpage

\bibliographystyle{unsrt}
\bibliography{thebib}

\end{document}